\documentclass{article}

\usepackage[preprint]{neurips_2025}

\usepackage[utf8]{inputenc} 
\usepackage[T1]{fontenc}    
\usepackage{hyperref}       
\usepackage{url}            
\usepackage{booktabs}       
\usepackage{amsfonts}       
\usepackage{nicefrac}       
\usepackage{microtype}      
\usepackage{xcolor}         

\usepackage{tabularx}
\usepackage{colortbl}
\usepackage{algorithm}
\usepackage{algorithmic}
\usepackage{wrapfig}
\usepackage{arydshln}
\usepackage{caption}
\usepackage[pdftex]{graphicx}
\usepackage{ulem}
\usepackage{enumitem}
\usepackage{multirow}
\usepackage{amsmath}

\definecolor{darkgreen}{RGB}{40,128,40}
\definecolor{darkblue}{RGB}{40,40,128}

\title{Think Before You Accept: Semantic Reflective Verification for Faster Speculative Decoding}

\author{%
  Yixuan Wang\quad{Yijun Liu}\quad{Shiyu Ji}\quad{Yuzhuang Xu}\\
  \textbf{Yang Xu}\quad\textbf{Qingfu Zhu}\quad\textbf{Wanxiang Che}\thanks{Corresponding author.} \\
  Research Center for Social Computing and Interactive Robotics, \\
  Harbin Institute of Technology, China \\
  \texttt{\{yixuanwang,car\}@ir.hit.edu.cn} \\
}

\begin{document}

\maketitle

\begin{abstract}
Large language models (LLMs) suffer from high inference latency due to the auto-regressive decoding process.
Speculative decoding accelerates inference by generating multiple draft tokens using a lightweight model
and verifying them in parallel.
However, existing verification methods rely heavily on distributional consistency while overlooking semantic correctness,
thereby limiting the potential speedup of speculative decoding.
While some methods employ additional models for relaxed verification of draft tokens,
they often fail to generalize effectively to more diverse or open-domain settings.
In this work, we propose Reflective Verification,
a training-free and semantics-aware approach that achieves a better trade-off between correctness and efficiency.
Specifically, we leverage the inherent reflective capacity of LLMs
to semantically assess the correctness of draft tokens in parallel during verification.
Using prompt-based probing, we obtain both the original and reflective distributions of draft tokens in a single forward pass.
The fusion of these distributions enables semantic-level verification of draft tokens
that incorporates both consistency and correctness.
Experiments across multiple domain benchmarks and model scales demonstrate that
our method significantly increases the acceptance length of draft tokens without compromising model performance.
Furthermore, we find that the proposed Reflective Verification is orthogonal to existing statistical verification methods,
and their combination yields additional 5$\sim$15\% improvements in decoding speed.
\end{abstract}

\section{Introduction}
Large language models (LLMs), such as ChatGPT \citep{achiam2023gpt} and Deepseek \citep{liu2024deepseek},
have demonstrated remarkable performance across a wide range of domains.
However, they also face numerous challenges \citep{zhou2024survey} during the deployment phase.
One major contributor to the high inference latency of LLMs is the auto-regressive decoding mechanism inherent in decoder-only architectures.
To mitigate the memory access bottlenecks associated with token-by-token generation, 
speculative decoding \citep{xia2024unlocking} has recently emerged as a promising approach for inference acceleration. 
This technique employs a lightweight draft model to propose multiple candidate tokens, which are then simultaneously verified by the target model.

Compared to the drafting stage, the primary objective of the verification stage is to determine whether the current candidate tokens are accepted.
Using exact match as the verification criterion \citep{xia2022speculative, santilli2023accelerating} can indeed ensure the losslessness of the acceleration method.
But it is often constrained in scenarios with high sampling temperatures.
In order to further enhance the acceleration effectiveness of speculative decoding,
several recent efforts \citep{kim2023speculative} have focused on developing more relaxed verification strategies
that increase the acceptance length of candidate tokens while preserving output correctness.
Various statistical metrics have been employed to enhance the reliability of verification strategies.
Speculative sampling \citep{leviathan2023fast,chen2023accelerating} proposes an unbiased decoding strategy with respect to the original distribution of the target LLM,
allowing flexible adjustment of the output based on the alignment between the draft and target distributions.
In addition, several studies have explored the use of neural networks to decide whether candidate drafts should be accepted.
Judge Decoding \citep{bachmann2025judge} trains a classifier on human-annotated data to achieve longer acceptance lengths without compromising downstream task performance.
\citet{liao2025reward} introduces an auxiliary reward model to evaluate drafts,
allowing for step-level discrimination of candidate tokens.

While the above methods improve the performance of speculative decoding during verification,
two key challenges remain unresolved.
\textbf{(1) The lack of semantic guidance.}
Current mainstream approaches primarily perform verification using statistical information between the draft and target distributions,
which provides limited information.
There is a need for a more efficient verification mechanism guided by semantic-level information,
where acceptance decisions are made based on semantic correctness rather than distributional consistency.
\textbf{(2) Limited generalization.}
Although some existing methods leverage deep models for verification,
they typically require additional human annotations and training procedures.
Moreover, many of these methods are tailored specifically for reasoning tasks with step-level responses,
and thus struggle to generalize to more general scenarios.

\begin{figure}[th]
  \centering
  \includegraphics[width=\textwidth]{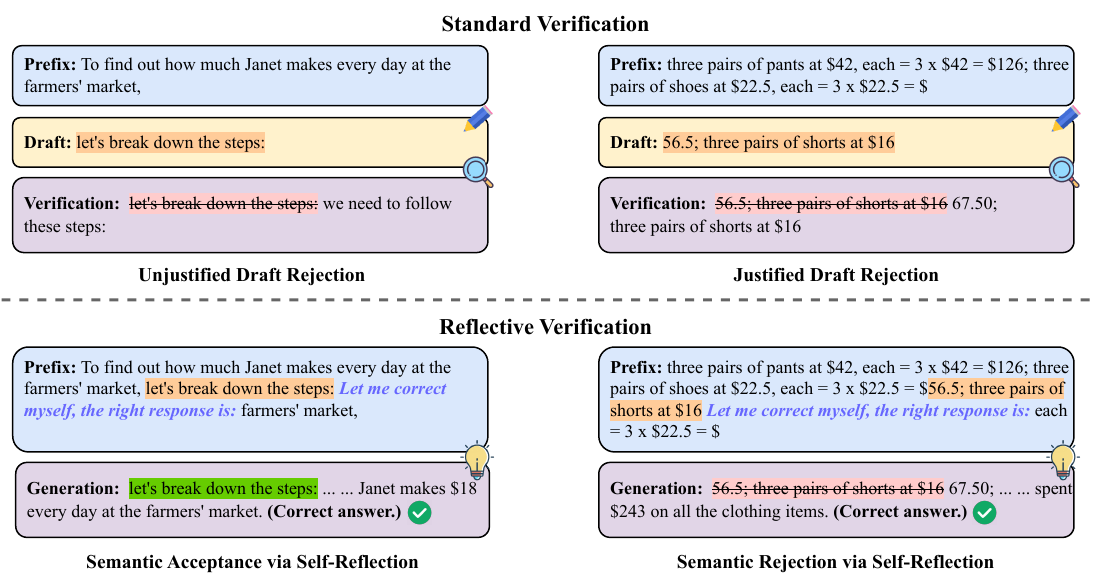}
  \caption{An illustration of draft tokens rejected by standard speculative decoding. Self-reflection enables the acceptance of semantically correct drafts that would otherwise be rejected.}
  \label{fig:reflect}
\end{figure}

In this paper, we propose \textbf{Reflective Verification}, a \textbf{training-free} draft verification method
that operates at the \textbf{semantic level} to address the aforementioned challenges.
Inspired by the observation in Figure \ref{fig:reflect} that self-reflection can effectively identify the semantic correctness of draft tokens,
we leverage prompt-based probing to explicitly trigger reflection of the model within a single forward pass.
The outcome of this reflection is then used to guide the verification of candidate draft tokens.
Specifically, we exploit the unidirectional attention mechanism of LLMs by appending a reflection prompt and a copy of the draft tokens after original draft tokens during verification.
This allows us to obtain the reflective judgment of target models on the current candidate tokens during the verification process.
By integrating the original output representing consistency with the reflective output ensuring correctness,
Reflective Verification can significantly extend the acceptance length of drafts while maintaining correctness,
thereby improving the speedup of speculative decoding.
In addition, the proposed method primarily calibrates the original output probabilities using reflective probabilities,
which is orthogonal to existing statistical-based verification mechanisms.
We conduct experiments across multiple configurations on benchmarks from various domains.
The results demonstrate that the proposed method can bring orthogonal improvements to a wide range of existing verification strategies,
achieving faster decoding without compromising task performance.
Moreover, under low-quality draft settings,
Reflective Verification helps mitigate the performance degradation of lossy verification methods
and can even lead to overall performance improvements.

Our main contributions are as follows:
\begin{itemize}[leftmargin=20pt]
    \item We present a plug-and-play speculative decoding verification approach that
incorporates semantic correctness by leveraging the reflective abilities of LLMs.
    \item By fusing the original and reflective outputs, the proposed method can be adapted to nearly all existing draft models and verification strategies,
demonstrating strong generalization capability.
    \item Extensive results show that the proposed method can significantly extend draft acceptance length without degrading model performance,
yielding a 5$\sim$15\% orthogonal improvement in end-to-end throughput.
\end{itemize}

\section{Observations}
In this section, we present several phenomena related to the verification of draft tokens that we have observed during the speculative decoding phase.
Motivated by these observations, we further propose the Reflective Verification method.

\subsection{Not All Rejected Drafts Are Incorrect}
\label{sec:2.1}
As discussed above, an effective verification mechanism requires a careful trade-off between correctness and inference efficiency.
With the continuous advancement of small language models \citep{xiao2024densing}, the quality of their draft tokens increases accordingly.
Relying solely on strict consistency for draft verification can significantly limit the upper bound of speedup achievable by speculative decoding.
In order to investigate potential improvements in verification strategies,
we conduct an analysis of the draft tokens rejected by the standard speculative decoding verification process.

Figure \ref{fig:reflect} presents several examples of rejected draft tokens identified by the standard verification mechanism.
We observe that some of these tokens, despite distributional inconsistencies, are semantically equivalent to the correct outputs.
Accepting such tokens would not compromise the overall correctness of the response
but can significantly improve the decoding speed.
For example, in the case of unjustified draft rejection shown in the figure,
although the token-level edit distance between the draft and the ground-truth output is large, the two sentences convey the same meaning.
An effective verification strategy should accept such semantically correct drafts, thereby further improving the upper bound of speculative decoding.

Given this observation, we believe that current draft verification strategies remain suboptimal 
and will play an increasingly important role with the ongoing development of draft models.
Developing a more relaxed and principled verification method is of great significance to the field of speculative decoding.
In this paper, we explore how to leverage the reflection of LLMs to achieve semantic-level correctness
rather than mere distributional consistency.

\subsection{Self-Reflection Enables Correctness Verification}
Although humans can naturally assess the correctness of draft tokens,
verification mechanisms that depend only on statistical information from the draft and target model distributions
often find it difficult to produce reliable decisions.
Recent efforts \citep{bachmann2025judge} have aimed to achieve semantic-level speculative decoding
by training classifiers using manually annotated draft acceptance labels.
This approach typically requires additional annotation and training,
and is difficult to quickly adapt to texts from other domains.
Accordingly, we seek to explore training-free approaches that utilize the inherent capabilities of
LLMs for semantic-level similarity verification.

Recently, self-reflection \citep{madaan2023self,ye2024physics,chen2025towards} has garnered significant attention as a key property of LLMs.
This behavior involves refining initially generated outputs by prompting the model itself through in-context learning (ICL).
Motivated by this insight, we attempt to leverage the reflective behavior of LLMs to verify the semantic-level correctness of draft candidates.
Specifically, we employ carefully designed prompts to induce the LLM
to perform reflection and regeneration on the two rejected drafts discussed in Section \ref{sec:2.1}.

As shown in Figure \ref{fig:reflect}, we are surprised to find that, with reflective prompting, 
the LLM is capable of effectively distinguishing between draft candidates at the semantic level.
In addition, unlike direct generation,
the reflection process resembles an error-correction procedure applied to the original draft candidates,
which aims to eliminate incorrect parts while preserving the original distribution as much as possible.
This property of reflection makes it particularly suitable for use in speculative decoding verification,
as it allows for the acceptance of a greater number of draft tokens while maintaining semantic correctness.
Inspired by the above observations,
we attempt to utilize the reflective output as auxiliary guidance to aid the original distribution
in making more reliable verification decisions.

\section{Reflective Verification}

\begin{figure}[t]
  \centering
  \includegraphics[width=\textwidth]{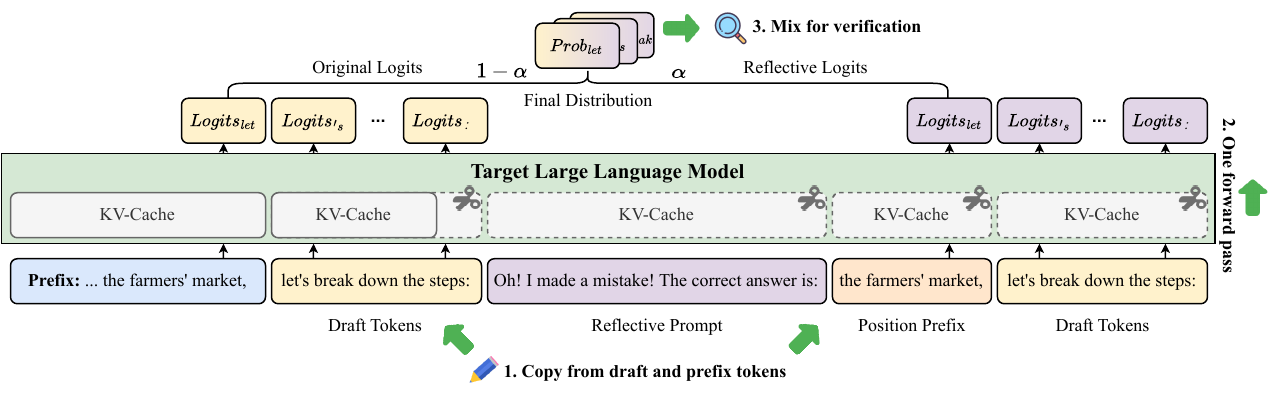}
  \caption{Overall structural diagram of Reflective Verification.
Compared to vanilla speculative decoding using only base outputs (yellow), we fuse them with reflective outputs (purple) as the final distribution.
}
  \label{fig:arch}
\end{figure}


\subsection{Extraction of Reflective Logits}
\label{sec:1}
The core of reflective verification lies in efficiently obtaining the reflection results of LLMs on draft candidates.
While standard self-reflection is capable of evaluating the draft tokens,
the reflection process itself still follows an auto-regressive decoding paradigm,
making it impossible for direct application in speculative decoding.
To address this, the proposed method employs prompt probing techniques to
obtain two output distributions over the draft tokens in a single forward pass.

As shown in Figure \ref{fig:arch}, we apply a specialized design to the original draft tokens
during the verification stage.
Instead of directly feeding the draft tokens, we maintain two identical copies of the draft,
with a reflective prompt probe inserted in between to explicitly trigger the reflection of LLMs.
Benefiting from the unidirectional attention mechanism, the subsequent template leaves the verification of the initial draft tokens unaffected.
The second draft tokens, informed by the context of probe, yields a reflection-based output that encodes semantic correctness verification.
Specifically, the draft sequence constructed at each step of speculative decoding can be formulated as:
\begin{equation}
     \mathrm{Draft_{final}} = \mathrm{Concat}(\mathrm{Draft_{ori}} \Vert  \mathrm{Prompt_{reflection}} \| \mathrm{Prefix_{position}} \| \mathrm{Draft_{ori}})
    \label{eq:prompt}
\end{equation}
where $\mathrm{Draft_{ori}}$ denotes the candidate tokens generated by the draft model,
$\mathrm{Prompt_{refection}}$ is a probe designed to prompt the model to reflect,
and $\mathrm{Prefix_{position}}$ refers to the tokens preceding the current draft candidate within the context,
serving to help the model locate the position for regeneration.

By feeding the constructed prompt into a single forward pass,
we can efficiently obtain two distinct distributions over the draft candidate.
Given the memory access bottlenecks during the decoding stage, the additional input does not significantly increase the forward latency.
It is important to note that, except for the first draft segment, the KV-cache entries associated with other parts do not participate in subsequent computations.
They are pruned after each forward pass, serving solely as a source of semantic-level verification signals.

\subsection{Fusion of Original and Reflective Logits}
\label{sec:2}
Despite sharing the same draft candidate tokens,
the logits produced by the LLM for each draft segment carry different interpretations.
The output of the first draft segment aligns with that of traditional speculative decoding
and represents the distribution at the consistency level.
By contrast, the second draft segment yields an output informed by the reflection of target LLM on the original draft,
capturing the distribution corresponding to semantic correctness.
To balance consistency with the original distribution and improved semantic correctness in the verification process,
we fuse the original logits and the reflective logits to form the final output distribution of the target LLM.
The final distribution for verification can be formulated as:
\begin{equation}
    Prob_{mix}[i] = \mathrm{Softmax}((1 - \alpha) * Logits[i] + \alpha * Logits[i+\mathrm{shift\_len}])
\end{equation}
We compute a weighted sum of the logits at position i and
its corresponding reflective logits to obtain the final output distribution.
$\mathrm{shift\_len}$ denotes the number of tokens occupied by the designed prompt and the first draft segment,
and $\alpha$ is a hyper-parameter that controls the weight of the reflective logits.

In essence, the proposed reflective verification mechanism uses the reflective logits as a side product
to selectively align the distribution of target model with the semantically correct distribution produced by draft model.
This method merely produces an output distribution with a higher acceptance rate and does not involve any specific verification mechanism.
Therefore, it is fully orthogonal to existing statistical verification approaches
and can be broadly applied across various draft models and verification settings.

\newcommand{\COMMENTLLAMA}[1]{{\color{darkgreen} $\triangleright$ {#1}}}
\newcommand{\COMMENTNEW}[1]{{\color{darkblue} $\triangleright$ {#1}}}

\begin{algorithm}[t]\footnotesize
    \caption{Speculative Sampling with Reflective Verification}
    \label{alg:main}
  \begin{algorithmic}
    \STATE {\bfseries Inputs:} $M_p, M_q, prefix, template, \alpha$.
    \FOR{$i=1$ {\bfseries to} $\gamma$}
      \STATE $q_i(x) \gets M_q(prefix + [x_1, \ldots, x_{i-1}])$
      \STATE $x_i \sim q_i(x)$
    \ENDFOR
    \STATE \COMMENTLLAMA{Prepare the reflective draft template.}
    \STATE $reflective\_draft \gets [x_1, \ldots, x_{\gamma}, template, x_1, \ldots, x_{\gamma}]$
    \STATE $m \gets \gamma + \vert template \vert + 1$
    \STATE \COMMENTLLAMA{Obtain the original logits $o_{1 : \gamma+1}(x)$ and reflective logits  $o_{m : m+\gamma}(x)$ in parallel.}
    \STATE $o_1(x), \ldots, o_{\gamma + 1}(x), o_{m}(x), \ldots, o_{m+\gamma}(x) \gets$ $M_p(prefix),\ldots , M_p(prefix + reflective\_draft)$
    \STATE \COMMENTLLAMA{Fuse the two logits to obtain the final distribution $p_{i}(x)$.}
    \STATE $p_1(x), \ldots, p_{\gamma + 1}(x) \gets softmax((1-\alpha)o_1(x)+\alpha o_{m}(x), \ldots, (1-\alpha)o_{\gamma + 1}(x)+\alpha o_{m+\gamma}(x))$
    \STATE $r_1 \sim U(0, 1), \dots, r_\gamma \sim U(0, 1)$
    \STATE $n \gets \min(\{ i - 1 \mid 1 \le i \le \gamma, r_i > \frac{p_i(x)}{q_i(x)} \} \cup \{ \gamma \})$
    \STATE $p'(x) \gets p_{n+1}(x)$
    \IF{$n < \gamma$}
      \STATE $p'(x) \gets norm(max(0, p_{n+1}(x) - q_{n+1}(x)))$
    \ENDIF
    \STATE $t \sim p'(x)$
    \STATE {\bfseries return} $prefix + [x_1, \ldots, x_{n}, t]$
  \end{algorithmic}
\end{algorithm}

\subsection{Speculative Decoding with Reflective Verification}
\label{sec:3}
Once the reflective output distribution is obtained,
the proposed method can be easily integrated with existing statistical verification approaches through minor modifications.
To further illustrate how the proposed method can be applied,
we take speculative sampling \citep{leviathan2023fast,chen2023accelerating} verification as an example to present the overall algorithmic process.
As demonstrated in Algorithm \ref{alg:main},
the parts marked with green annotations denote the primary distinctions between reflective verification and standard speculative sampling.
After obtaining the draft sequence, we construct the reflective draft for verification,
which includes two copies of the draft tokens and a reflection prompt.
By performing a single forward pass to compute the outputs of both copies in parallel and fusing them,
we obtain the probability $p_{i}(x)$ of the target LLM, corresponding to that in standard method.
Subsequently, all operations are identical to those in the standard speculative sampling verification mechanism.

It is worth noting that reflective process is fully decoupled from the draft generation and verification stages.
This also makes it compatible with nearly all draft generation and verification methods.
See Appendix \ref{alg:more} for details on the integration of Reflective Verification with other statistical methods.

\section{Experiments}
\subsection{Setting}

\paragraph{Benchmarks and metrics.}
In this paper, we select three commonly used benchmarks for different domains:
MT-Bench \citep{zheng2023judging} for dialogue, GSM8K \citep{cobbe2021training} for mathematics, and HumanEval \citep{chen2021evaluating} for code.
We use the corresponding metrics as task performance indicators,
with mean accepted tokens (\#MAT) \citep{xia2024unlocking} for each forward pass and end-to-end throughput serving as speed metrics.

\paragraph{Selected baselines.}
To demonstrate the generality of the proposed method,
we conduct experiments across multiple draft model configurations and verification strategies.
For the draft models, we choose two configurations from Llama3 series \citep{grattafiori2024llama} (1B\&8B and 8B\&70B)
to investigate the impact of model scale on reflective verification.

As for verification strategies, we select the following three commonly used methods:
(1) \textbf{Speculative Decoding} represents the naive lossless verification,
in which verification is deemed successful only when the draft tokens exactly match the sampling of the target model.
(2) \textbf{Speculative Sampling} \citep{leviathan2023fast} uses the probability ratio of candidate tokens under the target and draft distributions as the criterion,
achieving unbiased verification through sampling.
(3) \textbf{Typical Sampling} \citep{cai2024medusa} relaxes the verification criterion by 
using an entropy-based threshold derived from the target distribution,
significantly improving the acceptance rate of draft tokens.



\paragraph{Generation config.}
Since all models used in the experiments are instruct versions, we perform generation in a zero-shot manner across all three datasets.
Further details about hyperparameters, including $\alpha$, prefix length, and others, can be found in Appendix \ref{app:hyper}.
All experiments are conducted on a server equipped with two NVIDIA A100 GPUs (80GB each) and Intel(R) Xeon(R) Gold 6348 CPU @ 2.60GHz.


\subsection{Main Results}
The main experiments are conducted across diverse settings, with the detailed results presented in Table \ref{tab:main}.
We apply the proposed method (Reflec Verify) to various statistical verification approaches under two draft model settings,
and evaluate its impact on both task performance and acceleration performance.
Overall, Reflective Verification is orthogonal to existing common verification methods.
It can significantly increase the acceptance length of draft candidates, leading to improved end-to-end inference speed in speculative decoding.
Notably, this improvement comes without significant task performance degradation, and may even enhance it in some cases.
\paragraph{Acceleration performance.}
Initially, Reflective Verification consistently improves acceleration performance across various existing verification methods.
Under a fixed draft length, it yields an acceptance length increase close to 1, with particularly notable gains in mathematics and code generation tasks.
This leads to a 5$\sim$15\% orthogonal improvement in end-to-end throughput, achieved in a training-free and plug-and-play manner.
Moreover, the proposed method still brings improvements under typical sampling,
which already achieves the highest acceptance length,
demonstrating the broad applicability of our approach.

\begin{table}[t]\footnotesize
\centering
\caption{Main results across multiple benchmarks. \underline{Underline} denotes performance degradation, and O.T. denotes output tokens length.
\textbf{Bold} indicates the best result under each verification strategy.}
\label{tab:main}
\begin{tabularx}{\textwidth}{@{}l *{10}{>{\centering\arraybackslash}X}@{}}
\toprule
\multirow{2}{*}{\textbf{Method}} & \multicolumn{2}{c}{\textbf{MT-Bench}} & \multicolumn{2}{c}{\textbf{GSM8K}} & \multicolumn{2}{c}{\textbf{HumanEval}} & \multicolumn{4}{c}{\textbf{Average}} \\
 \cmidrule(l){2-3} 
 \cmidrule(l){4-5} 
 \cmidrule(l){6-7} 
 \cmidrule(l){8-11} 
 & Score & \#MAT & Acc. & \#MAT & Pass@1 & \#MAT & Perf. & O.T. & Tok./s & Speed \\
\midrule
\rowcolor{gray!15} \multicolumn{11}{c}{Llama3.2-1B-Instruct \& Llama3.1-8B-Instruct} \\
Vanilla AR & 7.44 & 1.00 & 77.63 & 1.00 & 65.85 & 1.00 & 72.63 & 477.52 & 45.96 & 1.00$\times$   \\
\hdashline
Spec Decoding & \textbf{7.44} & 3.43 & \textbf{77.63} & 6.12 & 65.85 & 6.68 & 72.63 & 480.88 & 51.02 & 1.11$\times$ \\
\textbf{+ Reflect Verify} & \underline{7.37}  & \textbf{4.15} & \underline{77.41} & \textbf{7.02} & \textbf{68.90} & \textbf{7.60} & \textbf{73.25} & 484.40 & \textbf{58.84} & \textbf{1.28$\times$} \\
\hdashline[0.5pt/2pt]
Spec Sampling & \textbf{7.51} & 4.03  & \textbf{78.09} & 6.25 & 66.46 & 6.77 & 73.05 & 479.20 & 54.13 & 1.18$\times$ \\
\textbf{+ Reflect Verify} & 7.44 & \textbf{4.88} & \textbf{78.09} & \textbf{7.15} & \textbf{69.51} & \textbf{7.65} & \textbf{73.92} & 472.67 & \textbf{62.32} & \textbf{1.36$\times$} \\
\hdashline[0.5pt/2pt]
Typical Sampling  & \textbf{7.65} & 4.82 & \underline{76.57} & 6.76 & \underline{63.41} & 7.27 & \underline{71.88} & 476.96 & 60.19 & 1.31$\times$ \\
\textbf{+ Reflect Verify} & 7.50 & \textbf{5.18} & \textbf{\underline{76.65}} & \textbf{7.50} & \textbf{67.68} & \textbf{7.93} & \textbf{73.00} & 485.93 & \textbf{64.79} & \textbf{1.41$\times$} \\
\midrule
\rowcolor{gray!15} \multicolumn{11}{c}{Llama3.1-8B-Instruct \& Llama3.1-70B-Instruct} \\
Vanilla AR & 8.24 & 1.00 & 84.91 & 1.00 &78.05 & 1.00 & 81.74 & 409.27 & 9.55 & 1.00$\times$ \\
\hdashline
Spec Decoding & 8.24 & 4.68 & 84.91 & 7.89 &78.05 & 8.48  & 81.74 & 408.59 & 18.09 & 1.89$\times$ \\
\textbf{+ Reflect Verify} & \textbf{8.34} & \textbf{5.93} & \textbf{85.06} & \textbf{9.33} & \textbf{78.66} & \textbf{9.52} & \textbf{82.33} & 411.63 & \textbf{20.00} & \textbf{2.09$\times$} \\
\hdashline[0.5pt/2pt]
Spec Sampling & 8.32 & 5.82 & 85.57 & 8.07 & 78.66 & 8.65 & 82.43 & 415.39 & 19.86 & 2.08$\times$ \\
\textbf{+ Reflect Verify} & \textbf{8.51} & \textbf{7.48} & \textbf{85.75} & \textbf{9.45} & \textbf{79.27} & \textbf{9.55} & \textbf{83.32} & 404.50 & \textbf{21.35} & \textbf{2.24$\times$}\\
\hdashline[0.5pt/2pt]
Typical Sampling  & \underline{7.93} & 7.24 & \textbf{85.52} & 8.81 & \underline{77.44} & 9.00  & \underline{80.68} & 410.11 & 21.44 & 2.25$\times$ \\
\textbf{+ Reflect Verify} & \textbf{\underline{8.17}} & \textbf{7.94} & 84.91 & \textbf{9.80} & \textbf{80.49} & \textbf{9.86} & \textbf{82.35} & 417.55 & \textbf{22.72} & \textbf{2.38$\times$}\\
\bottomrule
\end{tabularx}
\end{table}

\paragraph{Task performance.}
It is worth noting that Reflective Verification also brings certain improvements in task performance under lossy verification strategies.
Although existing lossy verification strategies significantly increase acceptance length, they often introduce degradation in generation quality,
which is particularly pronounced in objective tasks such as mathematics and code generation.
The core issue is that distributional statistics alone cannot ensure the semantic correctness of draft tokens, often resulting in the acceptance of incorrect drafts.
By leveraging the reflective signals of target LLMs, the proposed method enables semantic-level acceptance decisions for draft tokens.
As shown in the table, incorporating semantic information not only increases the acceptance length but also effectively mitigates performance degradation.
Moreover, incorporating Reflective Verification does not affect the overall output tokens length of target model.
We further provide case studies in Appendix \ref{app:case} to illustrate the semantic consistency brought by the proposed method.

\paragraph{Scale analysis.}
To investigate the impact of reflective capability on the proposed method, we also conduct experiments under two different draft model configurations.
While Reflective Verification provides improvements across different settings,
its impact on performance is particularly significant in the 70B target model configuration.
We attribute this to the fact that larger target LLMs possess stronger in-context learning and reflective capabilities,
enabling them to make more informed judgments based on the provided prompts.
This suggests that the proposed method holds great potential,
with its effectiveness expected to improve as the scale and capabilities of LLMs increase.

\begin{figure}[t]
\centering
\begin{minipage}{0.50\textwidth}\footnotesize
    \centering
    \includegraphics[width=\textwidth]{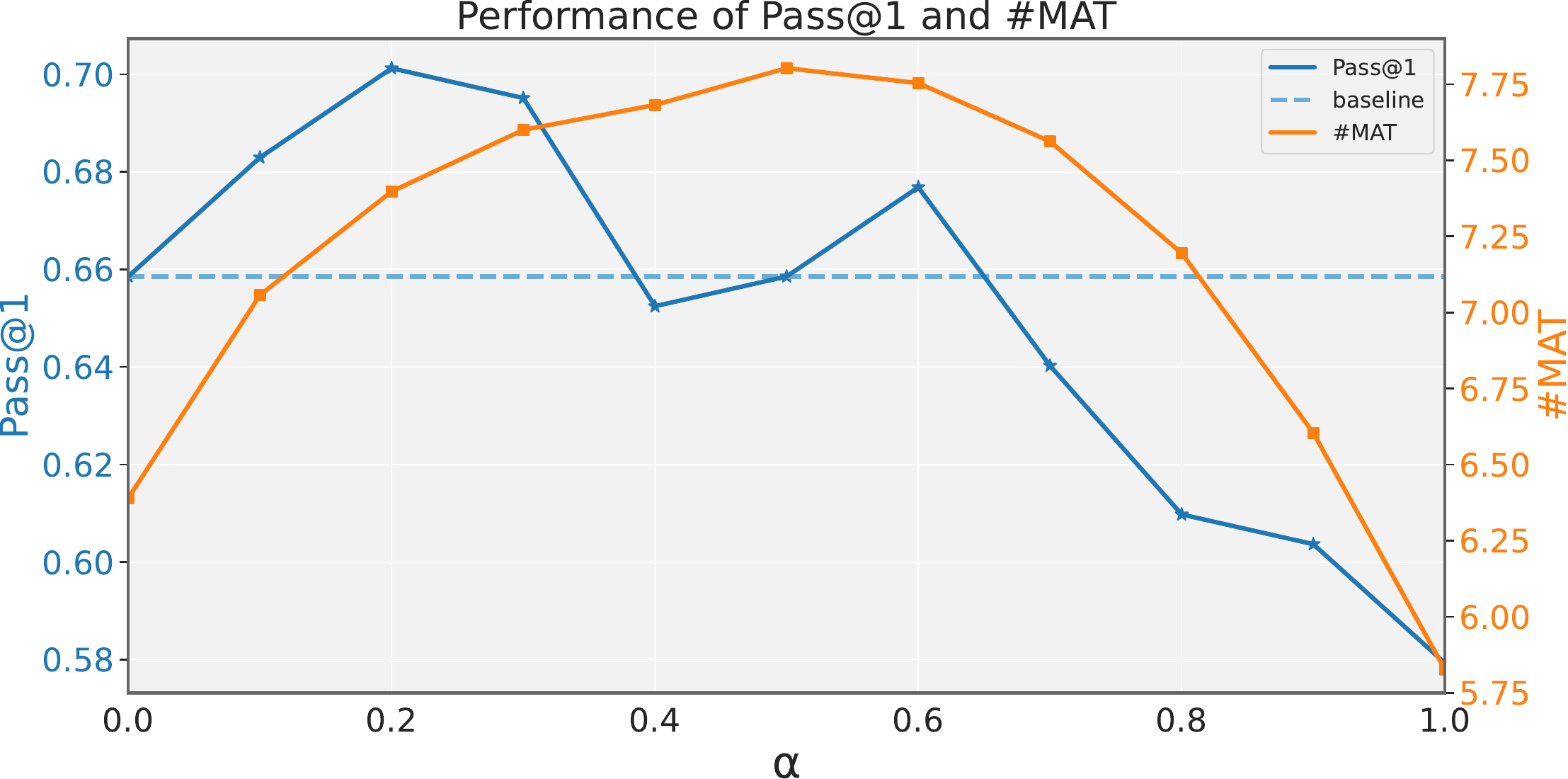}
    \captionof{figure}{Effect of $\alpha$ on task and acceleration performance.}
    \label{fig:alpha}
\end{minipage}
\hfill
\begin{minipage}{0.45\textwidth}\footnotesize
    \centering
    \includegraphics[width=\textwidth]{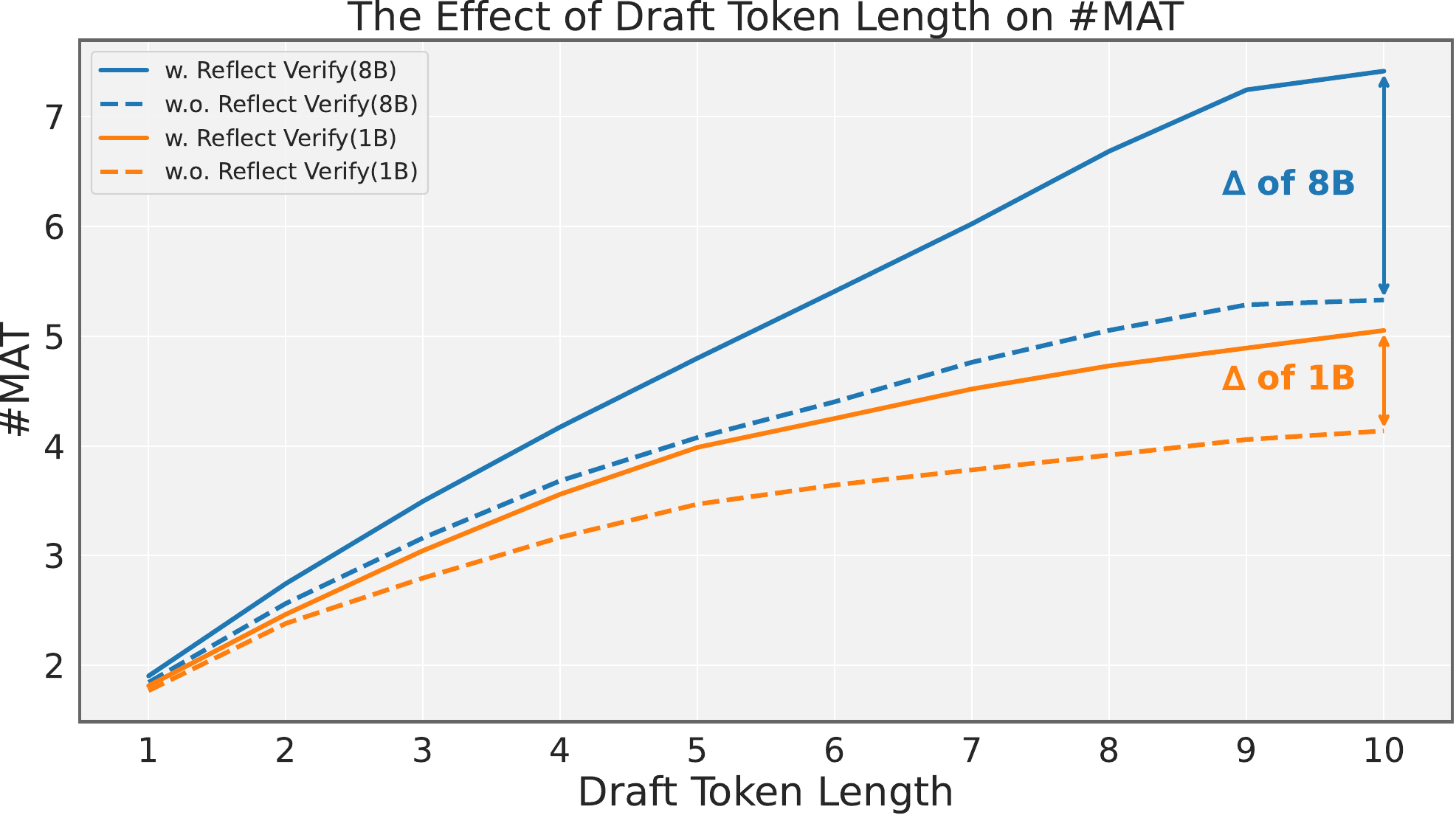}
    \captionof{figure}{Impact of draft quality on Reflective Verification.}
    \label{fig:quality}
\end{minipage}
\end{figure}

\section{Analysis}

\subsection{The Trade-off Impact of Alpha}
As a parameter controlling the weight of the reflective logits fusion,
$\alpha$ plays a crucial role in the performance of the reflective verification process.
To investigate how $\alpha$ balances consistency and semantic correctness in the reflective verification process,
we conduct an ablation study on the hyper-parameter.
As shown in Figure \ref{fig:alpha}, a trade-off relationship is observed between the value of $\alpha$  and the overall performance.
When $\alpha$  increases from a low value, the growing influence of reflective logits leads to consistent and significant improvements in task performance and the number of accepted tokens.
However, fully substituting the output distribution with reflective logits leads to performance degradation.
Since the method is training-free, the reflective ability of LLMs is not reliable enough to maintain consistency with the original distribution.
Based on the ablation results, we set the hyper-parameter to 0.3 in our experiments.

\subsection{Impact of Draft Quality}
As a verification approach, Reflective Verification does not directly enhance the quality of the draft itself.
Instead, it aims to maximize the acceptance rate of semantically correct drafts under a given set of candidate drafts.
Therefore, we conduct speculative decoding experiments using 1B and 8B draft models
to investigate the effectiveness of the proposed method under varying draft quality conditions.
Under the assumption that the 8B model generates higher-quality drafts,
we evaluate the improvements brought by Reflective Verification over traditional methods across different draft lengths on the MT-Bench dataset.

As shown in Figure \ref{fig:quality}, the proposed method, by incorporating semantic information,
significantly improves the acceptance rate at a fixed draft length,
thereby raising the upper bound of speculative decoding performance.
Notably, for the higher-quality drafts generated by the 8B model,
Reflective Verification yields even greater improvements.
This indicates that the verification mechanism does not merely increase the acceptance rate indiscriminately,
but rather makes informed decisions based on the semantic correctness of the draft.
In addition, with the advancement of draft models and improvements in draft quality,
this semantics-level verification approach is expected to exhibit even greater potential.

\subsection{Robustness of the Reflective Prompt}
Reflective Verification leverages constructed reflective prompts as probes to elicit reflective capabilities from LLMs.
The degree of sensitivity to these prompts directly influences the generalization capability of the proposed method.
To evaluate the robustness of the proposed method to reflective prompts,
we conduct experiments using a variety of alternative reflective prompts, as shown in Table \ref{tab:prompt}.
Specifically, the first row corresponds to standard speculative decoding without reflection, while the last row represents Reflective Verification with an empty reflective prompt.

It can be observed that the proposed method exhibits strong robustness to the reflective prompt.
Whether using a full sentence or a simple token such as [BACK],
it consistently outperforms standard speculative decoding.
This demonstrates that the process of obtaining reflective logits by duplicating the draft tokens does not rely heavily on prompt engineering,
allowing the method to be easily adapted to other settings.

\begin{table}[t]\footnotesize
    \caption{Robustness of reflective prompts. \underline{Underline} denotes performance degradation.}
    \label{tab:prompt}
    \begin{tabularx}{\textwidth}{Xcc}
    \toprule
    Reflection Prompt & Pass@1 & \#MAT \\
    \midrule
    \$\{draft\}  & 65.85 & 6.39 \\
    \hdashline
    \$\{draft\} Oh! I made a mistake! The correct answer is: \$\{prefix\} \$\{draft\} & \textbf{69.51} & \textbf{7.63} \\
    \$\{draft\} Let me correct myself, the right response is: \$\{prefix\} \$\{draft\} & 67.68 & 7.54 \\
    \$\{draft\} [BACK] \$\{prefix\} \$\{draft\} & 67.68 & 7.61 \\
    \$\{draft\} \$\{prefix\} \$\{draft\} & \underline{64.63} & 7.45 \\
    \bottomrule
  \end{tabularx}
\end{table}

\subsection{Comparison with Tree-Based Verification}

\begin{wraptable}{r}{0.5\textwidth}\footnotesize
\centering
\caption{A comparison with tree-based verification.}
\label{tab:Tree}
\begin{tabularx}{0.5\textwidth}{c*{1}{>{\centering\arraybackslash}X}cc}
    \toprule  
    \textbf{Method} & \textbf{Config} & \textbf{\#Budget} & \textbf{\#MAT} \\
    \midrule
    Chain & \{1x1x1x1x1\} & 5 & 3.08\\
    MCSD & \{4x2x2x1x1\} & 60 & 4.06 \\
    Ours & \{5+3+4+5\} & 17 & \textbf{4.92} \\
    \bottomrule
\end{tabularx}
\end{wraptable}

As a method that also leverages additional input tokens to improve acceptance rates,
tree-based verification \citep{miao2023specinfer} validates multiple candidate paths in a single forward pass by employing a sparse attention mask.
Under a fixed input budget, we compare our method with the representative approach MCSD \citep{yang2024multi} on the MT-Bench dataset.

The experimental results are shown in Table \ref{tab:Tree}. For MCSD, the configuration denotes the number of nodes at each tree depth.
For Reflective Verification, the configuration indicates the token counts of the four components in the prompt as defined in Equation \ref{eq:prompt}. The reflective prompt used is “[BACK]”.
Experimental results show that when the draft already yields a fluent output (e.g., 3.08 tokens accepted at chain mode.), the performance gains from traditional tree-based decoding become marginal.
Moreover, as the depth increases, the tree structure consumes a larger portion of the available budget.
In contrast, as draft generation models continue to improve,
the benefits of reflective verification are expected to increase, offering greater potential.


\section{Related Works}

\paragraph{Drafting methods of speculative decoding.}
As a core step in speculative decoding,
draft generation has attracted considerable attention from researchers \citep{xia2022speculative,bae2023fast,liu2023online,zhou2023distillspec}.
To obtain more consistent information, 
some methods \citep{stern2018blockwise,cai2024medusa} have begun leveraging the hidden states of the target model for draft prediction.
In particular, EAGLE \citep{li2024eagle} demonstrates significant acceleration by training a standalone single-layer transformer
designed to fuse token embeddings and the hidden states of the target model.

In contrast to approaches that rely on additional auxiliary models,
some methods \citep{yang2023inference,fu2024break,ou2024lossless,luo2024turning} aim to generate draft tokens more efficiently through retrieval-based techniques.
REST \citep{he2023rest} enables efficient draft tree construction and verification by building an index over the corpus.
In addition, some studies explore parallel decoding \citep{santilli2023accelerating} to
harness the capabilities of LLMs for self-drafting.
CLLMs \citep{kou2024cllms} improves the parallel decoding capability of LLMs by constructing and training on Jacobi decoding trajectories.
Despite differences in draft generation, acceptance rates consistently improve with Reflective Verification
by leveraging semantic signals.

\paragraph{Verification methods of speculative decoding.}
In addition to generating more consistent drafts,
numerous studies \citep{chen2023accelerating,leviathan2023fast} focus on improving verification methods to increase acceptance rates.
Under lossless acceleration, increasing acceptance rates hinges on the ability to verify multiple drafts simultaneously.
By utilizing sparse attention matrices, SpecInfer \citep{miao2023specinfer} accomplishes the verification of multiple draft paths in one forward computation,
leading to a notable improvement in acceptance length.
TR-Jacobi \citep{wang2024make} achieves an orthogonal fusion of model-based and retrieval-based methods
by incorporating retrieved paths into tree-based verification.

For lossy acceleration methods, the core lies in accepting as many inconsistent yet correct tokens as possible.
\citet{cai2024medusa} select plausible candidates for acceptance using an entropy-dependent threshold.
\citet{qin2024optimized} propose the multi-token joint decoding (MTJD), 
which performs verification based on the joint probability distribution rather than single token.
Although some methods \citep{bachmann2025judge,liao2025reward} leverage models with deep representations,
they typically require additional models and training.
We achieve a favorable balance between semantic-level validation and plug-and-play applicability.

\section{Discussion}
\paragraph{Broader impacts.}
The proposed method shifts the verification criterion from exact consistency to semantic correctness,
significantly raising the upper bound of speculative decoding and enabling the use of larger models as draft generators.
This may inspire more researchers to explore speculative decoding at the semantic level,
challenging the limitations of current paradigms.

\paragraph{Limitations \& Future.}
This work does not explore Reflective Verification on a wider range of draft models or larger-scale models (e.g., 405B).
While it significantly improves accepted draft length, it also increases step-wise variance, underscoring the need for dynamic draft length.
For fairness and control, we adopt a fixed draft length in this study, and leave its dynamic adaptation to future work.



\section{Conclusion}
In this paper, we introduce Reflective Verification, a training-free, semantic-level verification method for speculative decoding.
It is widely compatible with mainstream speculative decoding methods, boosting acceptance rates and enabling 5$\sim$15\% faster decoding with no performance degradation.
By incorporating semantic-level information,
the proposed method substantially expands the future potential of speculative decoding, especially as draft models continue to evolve.



\bibliographystyle{plainnat}
\bibliography{fix_references}

\begin{thebibliography}{35}
\providecommand{\natexlab}[1]{#1}
\providecommand{\url}[1]{\texttt{#1}}
\expandafter\ifx\csname urlstyle\endcsname\relax
  \providecommand{\doi}[1]{doi: #1}\else
  \providecommand{\doi}{doi: \begingroup \urlstyle{rm}\Url}\fi

\bibitem[Achiam et~al.(2023)Achiam, Adler, Agarwal, Ahmad, Akkaya, Aleman, Almeida, Altenschmidt, Altman, Anadkat, et~al.]{achiam2023gpt}
Josh Achiam, Steven Adler, Sandhini Agarwal, Lama Ahmad, Ilge Akkaya, Florencia~Leoni Aleman, Diogo Almeida, Janko Altenschmidt, Sam Altman, Shyamal Anadkat, et~al.
\newblock Gpt-4 technical report.
\newblock \emph{ArXiv preprint}, abs/2303.08774, 2023.
\newblock URL \url{https://arxiv.org/abs/2303.08774}.

\bibitem[Bachmann et~al.(2025)Bachmann, Anagnostidis, Pumarola, Georgopoulos, Sanakoyeu, Du, Sch{\"o}nfeld, Thabet, and Kohler]{bachmann2025judge}
Gregor Bachmann, Sotiris Anagnostidis, Albert Pumarola, Markos Georgopoulos, Artsiom Sanakoyeu, Yuming Du, Edgar Sch{\"o}nfeld, Ali Thabet, and Jonas Kohler.
\newblock Judge decoding: Faster speculative sampling requires going beyond model alignment.
\newblock \emph{ArXiv preprint}, abs/2501.19309, 2025.
\newblock URL \url{https://arxiv.org/abs/2501.19309}.

\bibitem[Bae et~al.(2023)Bae, Ko, Song, and Yun]{bae2023fast}
Sangmin Bae, Jongwoo Ko, Hwanjun Song, and Se-Young Yun.
\newblock Fast and robust early-exiting framework for autoregressive language models with synchronized parallel decoding.
\newblock In Houda Bouamor, Juan Pino, and Kalika Bali, editors, \emph{Proceedings of the 2023 Conference on Empirical Methods in Natural Language Processing}, pages 5910--5924, Singapore, 2023. Association for Computational Linguistics.
\newblock \doi{10.18653/v1/2023.emnlp-main.362}.
\newblock URL \url{https://aclanthology.org/2023.emnlp-main.362}.

\bibitem[Cai et~al.(2024)Cai, Li, Geng, Peng, Lee, Chen, and Dao]{cai2024medusa}
Tianle Cai, Yuhong Li, Zhengyang Geng, Hongwu Peng, Jason~D. Lee, Deming Chen, and Tri Dao.
\newblock Medusa: Simple {LLM} inference acceleration framework with multiple decoding heads.
\newblock In \emph{Forty-first International Conference on Machine Learning, {ICML} 2024, Vienna, Austria, July 21-27, 2024}. OpenReview.net, 2024.
\newblock URL \url{https://openreview.net/forum?id=PEpbUobfJv}.

\bibitem[Chen et~al.(2023)Chen, Borgeaud, Irving, Lespiau, Sifre, and Jumper]{chen2023accelerating}
Charlie Chen, Sebastian Borgeaud, Geoffrey Irving, Jean-Baptiste Lespiau, Laurent Sifre, and John Jumper.
\newblock Accelerating large language model decoding with speculative sampling.
\newblock \emph{ArXiv preprint}, abs/2302.01318, 2023.
\newblock URL \url{https://arxiv.org/abs/2302.01318}.

\bibitem[Chen et~al.(2021)Chen, Tworek, Jun, Yuan, Pinto, Kaplan, Edwards, Burda, Joseph, Brockman, et~al.]{chen2021evaluating}
Mark Chen, Jerry Tworek, Heewoo Jun, Qiming Yuan, Henrique Ponde De~Oliveira Pinto, Jared Kaplan, Harri Edwards, Yuri Burda, Nicholas Joseph, Greg Brockman, et~al.
\newblock Evaluating large language models trained on code.
\newblock \emph{ArXiv preprint}, abs/2107.03374, 2021.
\newblock URL \url{https://arxiv.org/abs/2107.03374}.

\bibitem[Chen et~al.(2025)Chen, Qin, Liu, Peng, Guan, Wang, Hu, Zhou, Gao, and Che]{chen2025towards}
Qiguang Chen, Libo Qin, Jinhao Liu, Dengyun Peng, Jiannan Guan, Peng Wang, Mengkang Hu, Yuhang Zhou, Te~Gao, and Wanxiang Che.
\newblock Towards reasoning era: A survey of long chain-of-thought for reasoning large language models.
\newblock \emph{ArXiv preprint}, abs/2503.09567, 2025.
\newblock URL \url{https://arxiv.org/abs/2503.09567}.

\bibitem[Cobbe et~al.(2021)Cobbe, Kosaraju, Bavarian, Chen, Jun, Kaiser, Plappert, Tworek, Hilton, Nakano, et~al.]{cobbe2021training}
Karl Cobbe, Vineet Kosaraju, Mohammad Bavarian, Mark Chen, Heewoo Jun, Lukasz Kaiser, Matthias Plappert, Jerry Tworek, Jacob Hilton, Reiichiro Nakano, et~al.
\newblock Training verifiers to solve math word problems.
\newblock \emph{ArXiv preprint}, abs/2110.14168, 2021.
\newblock URL \url{https://arxiv.org/abs/2110.14168}.

\bibitem[Fu et~al.(2024)Fu, Bailis, Stoica, and Zhang]{fu2024break}
Yichao Fu, Peter Bailis, Ion Stoica, and Hao Zhang.
\newblock Break the sequential dependency of {LLM} inference using lookahead decoding.
\newblock In \emph{Forty-first International Conference on Machine Learning, {ICML} 2024, Vienna, Austria, July 21-27, 2024}. OpenReview.net, 2024.
\newblock URL \url{https://openreview.net/forum?id=eDjvSFOkXw}.

\bibitem[Grattafiori et~al.(2024)Grattafiori, Dubey, Jauhri, Pandey, Kadian, Al-Dahle, Letman, Mathur, Schelten, Vaughan, et~al.]{grattafiori2024llama}
Aaron Grattafiori, Abhimanyu Dubey, Abhinav Jauhri, Abhinav Pandey, Abhishek Kadian, Ahmad Al-Dahle, Aiesha Letman, Akhil Mathur, Alan Schelten, Alex Vaughan, et~al.
\newblock The llama 3 herd of models.
\newblock \emph{ArXiv preprint}, abs/2407.21783, 2024.
\newblock URL \url{https://arxiv.org/abs/2407.21783}.

\bibitem[He et~al.(2024)He, Zhong, Cai, Lee, and He]{he2023rest}
Zhenyu He, Zexuan Zhong, Tianle Cai, Jason Lee, and Di~He.
\newblock {REST}: Retrieval-based speculative decoding.
\newblock In Kevin Duh, Helena Gomez, and Steven Bethard, editors, \emph{Proceedings of the 2024 Conference of the North American Chapter of the Association for Computational Linguistics: Human Language Technologies (Volume 1: Long Papers)}, pages 1582--1595, Mexico City, Mexico, 2024. Association for Computational Linguistics.
\newblock URL \url{https://aclanthology.org/2024.naacl-long.88}.

\bibitem[Kim et~al.(2023)Kim, Mangalam, Moon, Malik, Mahoney, Gholami, and Keutzer]{kim2023speculative}
Sehoon Kim, Karttikeya Mangalam, Suhong Moon, Jitendra Malik, Michael~W. Mahoney, Amir Gholami, and Kurt Keutzer.
\newblock Speculative decoding with big little decoder.
\newblock In Alice Oh, Tristan Naumann, Amir Globerson, Kate Saenko, Moritz Hardt, and Sergey Levine, editors, \emph{Advances in Neural Information Processing Systems 36: Annual Conference on Neural Information Processing Systems 2023, NeurIPS 2023, New Orleans, LA, USA, December 10 - 16, 2023}, 2023.
\newblock URL \url{http://papers.nips.cc/paper\_files/paper/2023/hash/7b97adeafa1c51cf65263459ca9d0d7c-Abstract-Conference.html}.

\bibitem[Kou et~al.(2024)Kou, Hu, He, Deng, and Zhang]{kou2024cllms}
Siqi Kou, Lanxiang Hu, Zhezhi He, Zhijie Deng, and Hao Zhang.
\newblock Cllms: Consistency large language models.
\newblock In \emph{Forty-first International Conference on Machine Learning, {ICML} 2024, Vienna, Austria, July 21-27, 2024}. OpenReview.net, 2024.
\newblock URL \url{https://openreview.net/forum?id=8uzBOVmh8H}.

\bibitem[Leviathan et~al.(2023)Leviathan, Kalman, and Matias]{leviathan2023fast}
Yaniv Leviathan, Matan Kalman, and Yossi Matias.
\newblock Fast inference from transformers via speculative decoding.
\newblock In Andreas Krause, Emma Brunskill, Kyunghyun Cho, Barbara Engelhardt, Sivan Sabato, and Jonathan Scarlett, editors, \emph{International Conference on Machine Learning, {ICML} 2023, 23-29 July 2023, Honolulu, Hawaii, {USA}}, volume 202 of \emph{Proceedings of Machine Learning Research}, pages 19274--19286. {PMLR}, 2023.
\newblock URL \url{https://proceedings.mlr.press/v202/leviathan23a.html}.

\bibitem[Li et~al.(2024)Li, Wei, Zhang, and Zhang]{li2024eagle}
Yuhui Li, Fangyun Wei, Chao Zhang, and Hongyang Zhang.
\newblock {EAGLE:} speculative sampling requires rethinking feature uncertainty.
\newblock In \emph{Forty-first International Conference on Machine Learning, {ICML} 2024, Vienna, Austria, July 21-27, 2024}. OpenReview.net, 2024.
\newblock URL \url{https://openreview.net/forum?id=1NdN7eXyb4}.

\bibitem[Liao et~al.(2025)Liao, Xu, Dong, Li, Monz, Savarese, Sahoo, and Xiong]{liao2025reward}
Baohao Liao, Yuhui Xu, Hanze Dong, Junnan Li, Christof Monz, Silvio Savarese, Doyen Sahoo, and Caiming Xiong.
\newblock Reward-guided speculative decoding for efficient llm reasoning.
\newblock \emph{ArXiv preprint}, abs/2501.19324, 2025.
\newblock URL \url{https://arxiv.org/abs/2501.19324}.

\bibitem[Liu et~al.(2024{\natexlab{a}})Liu, Feng, Xue, Wang, Wu, Lu, Zhao, Deng, Zhang, Ruan, et~al.]{liu2024deepseek}
Aixin Liu, Bei Feng, Bing Xue, Bingxuan Wang, Bochao Wu, Chengda Lu, Chenggang Zhao, Chengqi Deng, Chenyu Zhang, Chong Ruan, et~al.
\newblock Deepseek-v3 technical report.
\newblock \emph{ArXiv preprint}, abs/2412.19437, 2024{\natexlab{a}}.
\newblock URL \url{https://arxiv.org/abs/2412.19437}.

\bibitem[Liu et~al.(2024{\natexlab{b}})Liu, Hu, Bailis, Cheung, Deng, Stoica, and Zhang]{liu2023online}
Xiaoxuan Liu, Lanxiang Hu, Peter Bailis, Alvin Cheung, Zhijie Deng, Ion Stoica, and Hao Zhang.
\newblock Online speculative decoding.
\newblock In \emph{Forty-first International Conference on Machine Learning, {ICML} 2024, Vienna, Austria, July 21-27, 2024}. OpenReview.net, 2024{\natexlab{b}}.
\newblock URL \url{https://openreview.net/forum?id=BPQHXwVNvl}.

\bibitem[Luo et~al.(2024)Luo, Wang, Zhu, Zhang, Zhang, Yang, Xu, and Che]{luo2024turning}
Xianzhen Luo, Yixuan Wang, Qingfu Zhu, Zhiming Zhang, Xuanyu Zhang, Qing Yang, Dongliang Xu, and Wanxiang Che.
\newblock Turning trash into treasure: Accelerating inference of large language models with token recycling.
\newblock \emph{ArXiv preprint}, abs/2408.08696, 2024.
\newblock URL \url{https://arxiv.org/abs/2408.08696}.

\bibitem[Madaan et~al.(2023)Madaan, Tandon, Gupta, Hallinan, Gao, Wiegreffe, Alon, Dziri, Prabhumoye, Yang, Gupta, Majumder, Hermann, Welleck, Yazdanbakhsh, and Clark]{madaan2023self}
Aman Madaan, Niket Tandon, Prakhar Gupta, Skyler Hallinan, Luyu Gao, Sarah Wiegreffe, Uri Alon, Nouha Dziri, Shrimai Prabhumoye, Yiming Yang, Shashank Gupta, Bodhisattwa~Prasad Majumder, Katherine Hermann, Sean Welleck, Amir Yazdanbakhsh, and Peter Clark.
\newblock Self-refine: Iterative refinement with self-feedback.
\newblock In Alice Oh, Tristan Naumann, Amir Globerson, Kate Saenko, Moritz Hardt, and Sergey Levine, editors, \emph{Advances in Neural Information Processing Systems 36: Annual Conference on Neural Information Processing Systems 2023, NeurIPS 2023, New Orleans, LA, USA, December 10 - 16, 2023}, 2023.
\newblock URL \url{http://papers.nips.cc/paper\_files/paper/2023/hash/91edff07232fb1b55a505a9e9f6c0ff3-Abstract-Conference.html}.

\bibitem[Miao et~al.(2023)Miao, Oliaro, Zhang, Cheng, Wang, Wong, Chen, Arfeen, Abhyankar, and Jia]{miao2023specinfer}
Xupeng Miao, Gabriele Oliaro, Zhihao Zhang, Xinhao Cheng, Zeyu Wang, Rae Ying~Yee Wong, Zhuoming Chen, Daiyaan Arfeen, Reyna Abhyankar, and Zhihao Jia.
\newblock Specinfer: Accelerating generative llm serving with speculative inference and token tree verification.
\newblock \emph{ArXiv preprint}, abs/2305.09781, 2023.
\newblock URL \url{https://arxiv.org/abs/2305.09781}.

\bibitem[Ou et~al.(2024)Ou, Chen, and Tian]{ou2024lossless}
Jie Ou, Yueming Chen, and Prof. Tian.
\newblock Lossless acceleration of large language model via adaptive n-gram parallel decoding.
\newblock In Yi~Yang, Aida Davani, Avi Sil, and Anoop Kumar, editors, \emph{Proceedings of the 2024 Conference of the North American Chapter of the Association for Computational Linguistics: Human Language Technologies (Volume 6: Industry Track)}, pages 10--22, Mexico City, Mexico, 2024. Association for Computational Linguistics.
\newblock URL \url{https://aclanthology.org/2024.naacl-industry.2}.

\bibitem[Qin et~al.(2024)Qin, Hu, He, Prakriya, Cong, and Sun]{qin2024optimized}
Zongyue Qin, Ziniu Hu, Zifan He, Neha Prakriya, Jason Cong, and Yizhou Sun.
\newblock Optimized multi-token joint decoding with auxiliary model for llm inference.
\newblock \emph{ArXiv preprint}, abs/2407.09722, 2024.
\newblock URL \url{https://arxiv.org/abs/2407.09722}.

\bibitem[Santilli et~al.(2023)Santilli, Severino, Postolache, Maiorca, Mancusi, Marin, and Rodola]{santilli2023accelerating}
Andrea Santilli, Silvio Severino, Emilian Postolache, Valentino Maiorca, Michele Mancusi, Riccardo Marin, and Emanuele Rodola.
\newblock Accelerating transformer inference for translation via parallel decoding.
\newblock In Anna Rogers, Jordan Boyd-Graber, and Naoaki Okazaki, editors, \emph{Proceedings of the 61st Annual Meeting of the Association for Computational Linguistics (Volume 1: Long Papers)}, pages 12336--12355, Toronto, Canada, 2023. Association for Computational Linguistics.
\newblock \doi{10.18653/v1/2023.acl-long.689}.
\newblock URL \url{https://aclanthology.org/2023.acl-long.689}.

\bibitem[Stern et~al.(2018)Stern, Shazeer, and Uszkoreit]{stern2018blockwise}
Mitchell Stern, Noam Shazeer, and Jakob Uszkoreit.
\newblock Blockwise parallel decoding for deep autoregressive models.
\newblock In Samy Bengio, Hanna~M. Wallach, Hugo Larochelle, Kristen Grauman, Nicol{\`{o}} Cesa{-}Bianchi, and Roman Garnett, editors, \emph{Advances in Neural Information Processing Systems 31: Annual Conference on Neural Information Processing Systems 2018, NeurIPS 2018, December 3-8, 2018, Montr{\'{e}}al, Canada}, pages 10107--10116, 2018.
\newblock URL \url{https://proceedings.neurips.cc/paper/2018/hash/c4127b9194fe8562c64dc0f5bf2c93bc-Abstract.html}.

\bibitem[Wang et~al.(2024)Wang, Luo, Wei, Liu, Zhu, Zhang, Yang, Xu, and Che]{wang2024make}
Yixuan Wang, Xianzhen Luo, Fuxuan Wei, Yijun Liu, Qingfu Zhu, Xuanyu Zhang, Qing Yang, Dongliang Xu, and Wanxiang Che.
\newblock Make some noise: Unlocking language model parallel inference capability through noisy training.
\newblock \emph{ArXiv preprint}, abs/2406.17404, 2024.
\newblock URL \url{https://arxiv.org/abs/2406.17404}.

\bibitem[Xia et~al.(2023)Xia, Ge, Wang, Chen, Wei, and Sui]{xia2022speculative}
Heming Xia, Tao Ge, Peiyi Wang, Si-Qing Chen, Furu Wei, and Zhifang Sui.
\newblock Speculative decoding: Exploiting speculative execution for accelerating seq2seq generation.
\newblock In Houda Bouamor, Juan Pino, and Kalika Bali, editors, \emph{Findings of the Association for Computational Linguistics: EMNLP 2023}, pages 3909--3925, Singapore, 2023. Association for Computational Linguistics.
\newblock \doi{10.18653/v1/2023.findings-emnlp.257}.
\newblock URL \url{https://aclanthology.org/2023.findings-emnlp.257}.

\bibitem[Xia et~al.(2024)Xia, Yang, Dong, Wang, Li, Ge, Liu, Li, and Sui]{xia2024unlocking}
Heming Xia, Zhe Yang, Qingxiu Dong, Peiyi Wang, Yongqi Li, Tao Ge, Tianyu Liu, Wenjie Li, and Zhifang Sui.
\newblock Unlocking efficiency in large language model inference: A comprehensive survey of speculative decoding.
\newblock \emph{ArXiv preprint}, abs/2401.07851, 2024.
\newblock URL \url{https://arxiv.org/abs/2401.07851}.

\bibitem[Xiao et~al.(2024)Xiao, Cai, Zhao, Zeng, Lin, Zhou, Zheng, Han, Liu, and Sun]{xiao2024densing}
Chaojun Xiao, Jie Cai, Weilin Zhao, Guoyang Zeng, Biyuan Lin, Jie Zhou, Zhi Zheng, Xu~Han, Zhiyuan Liu, and Maosong Sun.
\newblock Densing law of llms.
\newblock \emph{ArXiv preprint}, abs/2412.04315, 2024.
\newblock URL \url{https://arxiv.org/abs/2412.04315}.

\bibitem[Yang et~al.(2023)Yang, Ge, Wang, Jiao, Jiang, Yang, Majumder, and Wei]{yang2023inference}
Nan Yang, Tao Ge, Liang Wang, Binxing Jiao, Daxin Jiang, Linjun Yang, Rangan Majumder, and Furu Wei.
\newblock Inference with reference: Lossless acceleration of large language models.
\newblock \emph{ArXiv preprint}, abs/2304.04487, 2023.
\newblock URL \url{https://arxiv.org/abs/2304.04487}.

\bibitem[Yang et~al.(2024)Yang, Huang, Dai, and Chen]{yang2024multi}
Sen Yang, Shujian Huang, Xinyu Dai, and Jiajun Chen.
\newblock Multi-candidate speculative decoding.
\newblock \emph{ArXiv preprint}, abs/2401.06706, 2024.
\newblock URL \url{https://arxiv.org/abs/2401.06706}.

\bibitem[Ye et~al.(2024)Ye, Xu, Li, and Allen-Zhu]{ye2024physics}
Tian Ye, Zicheng Xu, Yuanzhi Li, and Zeyuan Allen-Zhu.
\newblock Physics of language models: Part 2.2, how to learn from mistakes on grade-school math problems.
\newblock \emph{ArXiv preprint}, abs/2408.16293, 2024.
\newblock URL \url{https://arxiv.org/abs/2408.16293}.

\bibitem[Zheng et~al.(2023)Zheng, Chiang, Sheng, Zhuang, Wu, Zhuang, Lin, Li, Li, Xing, Zhang, Gonzalez, and Stoica]{zheng2023judging}
Lianmin Zheng, Wei{-}Lin Chiang, Ying Sheng, Siyuan Zhuang, Zhanghao Wu, Yonghao Zhuang, Zi~Lin, Zhuohan Li, Dacheng Li, Eric~P. Xing, Hao Zhang, Joseph~E. Gonzalez, and Ion Stoica.
\newblock Judging llm-as-a-judge with mt-bench and chatbot arena.
\newblock In Alice Oh, Tristan Naumann, Amir Globerson, Kate Saenko, Moritz Hardt, and Sergey Levine, editors, \emph{Advances in Neural Information Processing Systems 36: Annual Conference on Neural Information Processing Systems 2023, NeurIPS 2023, New Orleans, LA, USA, December 10 - 16, 2023}, 2023.
\newblock URL \url{http://papers.nips.cc/paper\_files/paper/2023/hash/91f18a1287b398d378ef22505bf41832-Abstract-Datasets\_and\_Benchmarks.html}.

\bibitem[Zhou et~al.(2024{\natexlab{a}})Zhou, Lyu, Rawat, Menon, Rostamizadeh, Kumar, Kagy, and Agarwal]{zhou2023distillspec}
Yongchao Zhou, Kaifeng Lyu, Ankit~Singh Rawat, Aditya~Krishna Menon, Afshin Rostamizadeh, Sanjiv Kumar, Jean{-}Fran{\c{c}}ois Kagy, and Rishabh Agarwal.
\newblock Distillspec: Improving speculative decoding via knowledge distillation.
\newblock In \emph{The Twelfth International Conference on Learning Representations, {ICLR} 2024, Vienna, Austria, May 7-11, 2024}. OpenReview.net, 2024{\natexlab{a}}.
\newblock URL \url{https://openreview.net/forum?id=rsY6J3ZaTF}.

\bibitem[Zhou et~al.(2024{\natexlab{b}})Zhou, Ning, Hong, Fu, Xu, Li, Lou, Wang, Yuan, Li, et~al.]{zhou2024survey}
Zixuan Zhou, Xuefei Ning, Ke~Hong, Tianyu Fu, Jiaming Xu, Shiyao Li, Yuming Lou, Luning Wang, Zhihang Yuan, Xiuhong Li, et~al.
\newblock A survey on efficient inference for large language models.
\newblock \emph{ArXiv preprint}, abs/2404.14294, 2024{\natexlab{b}}.
\newblock URL \url{https://arxiv.org/abs/2404.14294}.

\end{thebibliography}

\appendix

\section{Details in the Algorithm}
\label{alg:more}
We present the pseudocode of the proposed Reflective Verification integrated with speculative decoding and the typical sampling algorithm.
As shown in Algorithms \ref{alg:decoding} and \ref{alg:typical},
only minimal modifications are required to integrate Reflective Verification with existing statistics-based verification methods.

\begin{algorithm}[h]\footnotesize
    \caption{Speculative Decoding with Reflective Verification}
    \label{alg:decoding}
  \begin{algorithmic}
    \STATE {\bfseries Inputs:} $M_p, M_q, prefix, template, \alpha$.
    \FOR{$i=1$ {\bfseries to} $\gamma$}
      \STATE $q_i(x) \gets M_q(prefix + [x_1, \ldots, x_{i-1}])$
      \STATE $x_i \sim q_i(x)$
    \ENDFOR
    \STATE \COMMENTLLAMA{Prepare the reflective draft template.}
    \STATE $reflective\_draft \gets [x_1, \ldots, x_{\gamma}, template, x_1, \ldots, x_{\gamma}]$
    \STATE $m \gets \gamma + \vert template \vert + 1$
    \STATE \COMMENTLLAMA{Obtain the original logits $o_{1 : \gamma+1}(x)$ and reflective logits  $o_{m : m+\gamma}(x)$ in parallel.}
    \STATE $o_1(x), \ldots, o_{\gamma + 1}(x), o_{m}(x), \ldots, o_{m+\gamma}(x) \gets$ $M_p(prefix),\ldots , M_p(prefix + reflective\_draft)$
    \STATE \COMMENTLLAMA{Fuse the two logits to obtain the final distribution $p_{i}(x)$.}
    \STATE $p_1(x), \ldots, p_{\gamma + 1}(x) \gets softmax((1-\alpha)o_1(x)+\alpha o_{m}(x), \ldots, (1-\alpha)o_{\gamma + 1}(x)+\alpha o_{m+\gamma}(x))$
    \STATE \COMMENTNEW{Use exact match verification.}
    \STATE $\hat{x}_1 \sim p_1, \dots, \hat{x}_\gamma \sim p_{\gamma}$

    \STATE $n \gets \min(\{ i - 1 \mid 1 \le i \le \gamma, x_{i}=\hat{x}_i \} \cup \{ \gamma \})$
    \STATE $t \sim p_{n+1}(x)$
    \STATE {\bfseries return} $prefix + [x_1, \ldots, x_{n}, t]$
  \end{algorithmic}
\end{algorithm}

\begin{algorithm}[h]\footnotesize
  \caption{Typical Sampling with Reflective Verification}
  \label{alg:typical}
  \begin{algorithmic}
    \STATE {\bfseries Inputs:} $M_p, M_q, prefix, template, \alpha, \epsilon, \delta$.
    \FOR{$i=1$ {\bfseries to} $\gamma$}
      \STATE $q_i(x) \gets M_q(prefix + [x_1, \ldots, x_{i-1}])$
      \STATE $x_i \sim q_i(x)$
    \ENDFOR
    \STATE $reflective\_draft \gets [x_1, \ldots, x_{\gamma}, template, x_1, \ldots, x_{\gamma}]$
    \STATE $m \gets \gamma + \vert template \vert + 1$
    \STATE \COMMENTLLAMA{Obtain the original logits $o_{1 : \gamma+1}(x)$ and reflective logits  $o_{m : m+\gamma}(x)$ in parallel.}
    \STATE $o_1(x), \ldots, o_{\gamma + 1}(x), o_{m}(x), \ldots, o_{m+\gamma}(x) \gets$ $M_p(prefix),\ldots , M_p(prefix + reflective\_draft)$
    \STATE \COMMENTLLAMA{Fuse the two logits to obtain the final distribution $p_{i}(x)$.}
    \STATE $p_1(x), \ldots, p_{\gamma + 1}(x) \gets softmax((1-\alpha)o_1(x)+\alpha o_{m}(x), \ldots, (1-\alpha)o_{\gamma + 1}(x)+\alpha o_{m+\gamma}(x))$
    \STATE \COMMENTNEW{Use entropy-based threshold verification.}
    \STATE $ threshold = \min\left( \epsilon, \delta \exp\left( -H\left(p_{\text{original}}(\cdot \mid x_1, x_2, \cdots, x_{n+k-1})\right) \right) \right)$
    \STATE $n \gets \min(\{ i - 1 \mid 1 \le i \le \gamma, p_{i}(x) > threshold \} \cup \{ \gamma \})$
    \STATE $t \sim p_{n+1}(x)$
    \STATE {\bfseries return} $prefix + [x_1, \ldots, x_{n}, t]$
  \end{algorithmic}
\end{algorithm}

\newpage

\section{Hyperparameter Details}
\label{app:hyper}
For the main experiments in Table \ref{tab:main}, we adopt the following hyperparameter settings.
We present the configurations of speculative decoding and generation settings in Table \ref{tab:hyper}.
For different draft models and datasets, we select the optimal draft length K and temperature.

\begin{table}[h]\footnotesize
    \centering
    \caption{Details of the hyperparameters under different experimental settings.}
    \label{tab:hyper}
    \begin{tabularx}{\textwidth}{*{6}{>{\centering\arraybackslash}X}}
    \toprule
    \textbf{Setting} & \textbf{Dataset} & \textbf{Assistant K} & \boldmath{$\alpha$} & \textbf{Temperature} & \textbf{Prefix Len}\\
    \midrule
    1B\&8B & MT-Bench & 5 & 0.3 & 0.8 & 4 \\
    1B\&8B & GSM8K & 8 & 0.3 & 0.2 & 4 \\
    1B\&8B & HumanEval & 8 & 0.3 & 0.2 & 4 \\
    8B\&70B & MT-Bench & 8 & 0.3 & 0.8 & 4 \\
    8B\&70B & GSM8K & 10 & 0.3 & 0.2 & 4 \\
    8B\&70B & HumanEval & 10 & 0.3 & 0.2 & 4 \\
    \bottomrule
    \end{tabularx}
\end{table}

\section{Detailed Case Study}
\label{app:case}
\begin{figure}[h]
    \centering
    \includegraphics[width=\textwidth]{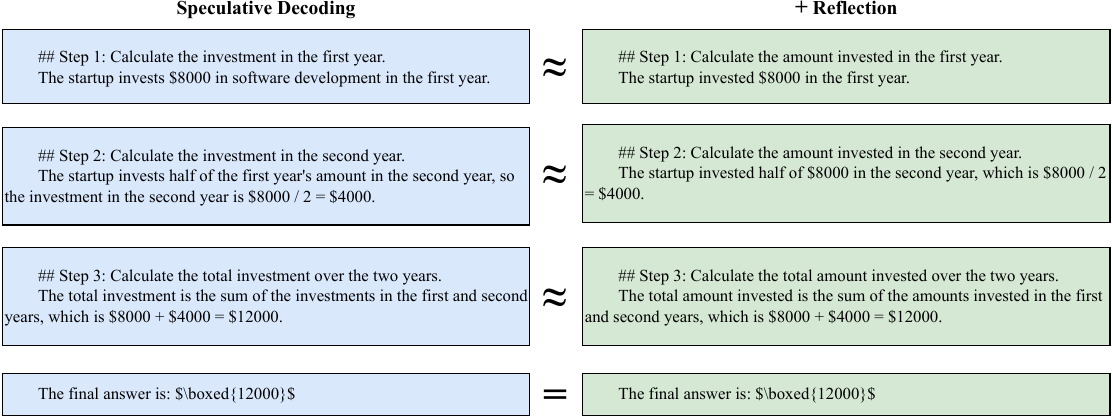}
    \caption{An illustration of reflective verification on MT-Bench.}
    \label{fig:case1}
\end{figure}

\begin{figure}[h]
    \centering
    \includegraphics[width=\textwidth]{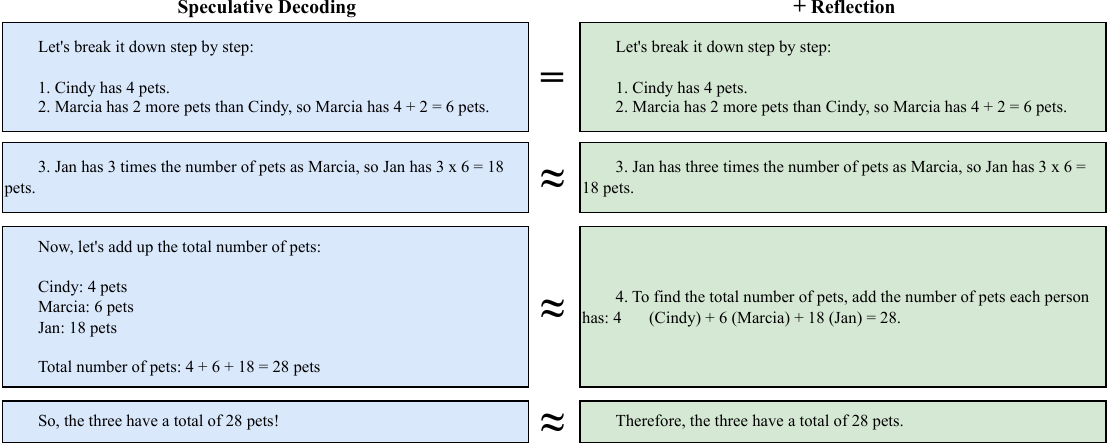}
    \caption{An illustration of reflective verification on GSM8K.}
    \label{fig:case2}
\end{figure}

To further investigate the impact of Reflective Verification on model outputs, we conduct case studies on three datasets and present several representative examples.
As shown in Figures \ref{fig:case1} and \ref{fig:case2}, we segment semantically similar blocks between the two outputs. Despite variations in phrasing, the overall output length and semantic content remain consistent.

\begin{figure}[h]
    \centering
    \includegraphics[width=\textwidth]{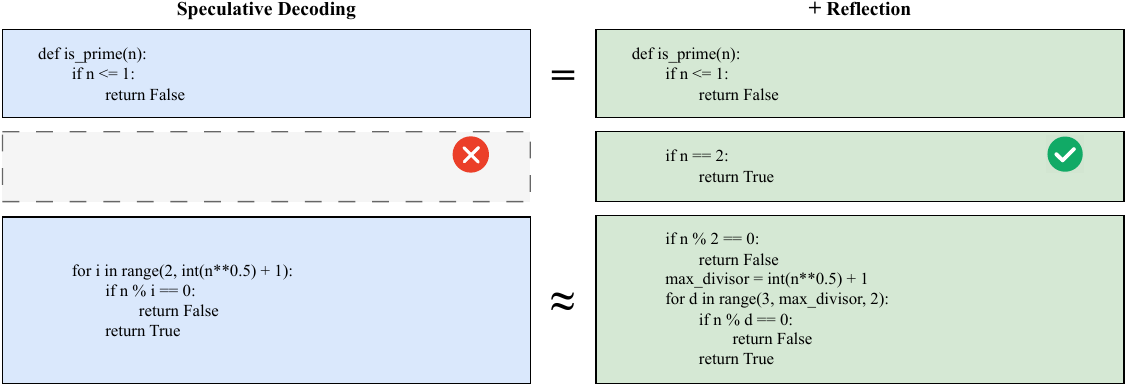}
    \caption{An illustration of reflective verification on HumanEval.}
    \label{fig:case3}
\end{figure}

Notably, in the code generation domain,
we observe that the performance gains from Reflective Verification primarily stem from improved handling of boundary cases.
As shown in Figure \ref{fig:case3}, through self-reflection, the model becomes more sensitive to such edge conditions.
By integrating reflective logits, it is able to generate higher-quality code.

\clearpage
\newpage
\section*{NeurIPS Paper Checklist}

\begin{enumerate}

\item {\bf Claims}
    \item[] Question: Do the main claims made in the abstract and introduction accurately reflect the paper's contributions and scope?
    \item[] Answer: \answerYes{} 
    \item[] Justification: Yes, we have outlined the main contributions and scope of the paper at the end of both the abstract and the introduction.

\item {\bf Limitations}
    \item[] Question: Does the paper discuss the limitations of the work performed by the authors?
    \item[] Answer: \answerYes{} 
    \item[] Justification: Yes, we briefly discuss the limitations of our work and potential future directions in Section 7.

\item {\bf Theory assumptions and proofs}
    \item[] Question: For each theoretical result, does the paper provide the full set of assumptions and a complete (and correct) proof?
    \item[] Answer: \answerNA{} 
    \item[] Justification: This paper does not include theoretical results.

    \item {\bf Experimental result reproducibility}
    \item[] Question: Does the paper fully disclose all the information needed to reproduce the main experimental results of the paper to the extent that it affects the main claims and/or conclusions of the paper (regardless of whether the code and data are provided or not)?
    \item[] Answer: \answerYes{} 
    \item[] Justification: Yes, we fully report our experimental settings in Section 4.1 and the appendix B to ensure reproducibility. Additionally, the algorithmic procedures involved are detailed in Section 3.3 and the appendix A.

\item {\bf Open access to data and code}
    \item[] Question: Does the paper provide open access to the data and code, with sufficient instructions to faithfully reproduce the main experimental results, as described in supplemental material?
    \item[] Answer: \answerNo{} 
    \item[] Justification: We will release our code in the official version of the subsequent paper to provide reproduction and provide more help to the future community.

\item {\bf Experimental setting/details}
    \item[] Question: Does the paper specify all the training and test details (e.g., data splits, hyperparameters, how they were chosen, type of optimizer, etc.) necessary to understand the results?
    \item[] Answer: \answerYes{} 
    \item[] Justification: Yes, we fully report our experimental settings in Section 4.1 and the appendix B to ensure reproducibility.

\item {\bf Experiment statistical significance}
    \item[] Question: Does the paper report error bars suitably and correctly defined or other appropriate information about the statistical significance of the experiments?
    \item[] Answer: \answerYes{} 
    \item[] Justification: In this work, we report task performance and speedup by averaging the sampling results over three runs.

\item {\bf Experiments compute resources}
    \item[] Question: For each experiment, does the paper provide sufficient information on the computer resources (type of compute workers, memory, time of execution) needed to reproduce the experiments?
    \item[] Answer: \answerYes{} 
    \item[] Justification: Yes, we describe the hardware environment used in our experiments in Section 4.1.
    
\item {\bf Code of ethics}
    \item[] Question: Does the research conducted in the paper conform, in every respect, with the NeurIPS Code of Ethics \url{https://neurips.cc/public/EthicsGuidelines}?
    \item[] Answer: \answerYes{} 
    \item[] Justification: Yes, we adhere to the NeurIPS Code of Ethics.

\item {\bf Broader impacts}
    \item[] Question: Does the paper discuss both potential positive societal impacts and negative societal impacts of the work performed?
    \item[] Answer: \answerYes{} 
    \item[] Justification: Yes, we discuss the broader impacts in Section 7.
    
\item {\bf Safeguards}
    \item[] Question: Does the paper describe safeguards that have been put in place for responsible release of data or models that have a high risk for misuse (e.g., pretrained language models, image generators, or scraped datasets)?
    \item[] Answer: \answerNA{} 
    \item[] Justification: This paper poses no such risks.

\item {\bf Licenses for existing assets}
    \item[] Question: Are the creators or original owners of assets (e.g., code, data, models), used in the paper, properly credited and are the license and terms of use explicitly mentioned and properly respected?
    \item[] Answer: \answerYes{} 
    \item[] Justification: Yes, we properly cite all the involved open-source models.

\item {\bf New assets}
    \item[] Question: Are new assets introduced in the paper well documented and is the documentation provided alongside the assets?
    \item[] Answer: \answerNA{} 
    \item[] Justification: The paper does not release new assets.

\item {\bf Crowdsourcing and research with human subjects}
    \item[] Question: For crowdsourcing experiments and research with human subjects, does the paper include the full text of instructions given to participants and screenshots, if applicable, as well as details about compensation (if any)? 
    \item[] Answer: \answerNA{} 
    \item[] Justification: The paper does not involve crowdsourcing nor research with human subjects.

\item {\bf Institutional review board (IRB) approvals or equivalent for research with human subjects}
    \item[] Question: Does the paper describe potential risks incurred by study participants, whether such risks were disclosed to the subjects, and whether Institutional Review Board (IRB) approvals (or an equivalent approval/review based on the requirements of your country or institution) were obtained?
    \item[] Answer: \answerNA{} 
    \item[] Justification: The paper does not involve crowdsourcing nor research with human subjects.

\item {\bf Declaration of LLM usage}
    \item[] Question: Does the paper describe the usage of LLMs if it is an important, original, or non-standard component of the core methods in this research? Note that if the LLM is used only for writing, editing, or formatting purposes and does not impact the core methodology, scientific rigorousness, or originality of the research, declaration is not required.
    \item[] Answer: \answerNA{} 
    \item[] Justification: LLMs are used only for some polishing tasks in the paper.

\end{enumerate}

\end{document}